\documentclass[10pt, conference, compsocconf]{IEEEtran}
\usepackage{cite}
\ifCLASSINFOpdf
\else
\fi
\usepackage[cmex10]{amsmath}
\usepackage{algorithmic}
\usepackage{amssymb,amsfonts}
\usepackage{graphicx}
\usepackage{textcomp}
\usepackage{xcolor}
\usepackage{mathtools}
\usepackage{standalone}
\usepackage{tikz}
\usepackage{bm}
\usepackage{siunitx}
\usepackage{hyperref}

\newcommand{\email}[1]{\href{mailto:#1}{\textit{#1}}}

\newcommand{\Real}{\mathbb R}

\newcommand{\isGaussian}[2]{\sim \mathcal{N}\left(#1,#2\right)}
\newcommand{\isGaussianBigg}[2]{\sim \mathcal{N}\Bigg(#1,#2\Bigg)}
\newcommand{\tp}[2][-6]{{#2}^{\mkern#1mu T}} % 'transpose' macro

\newcommand{\bcf}{\;\mbox{\boldmath ${\cal F}$\unboldmath}}
\newcommand{\bcfa}{\underrightarrow{\bcf}}
\def\Vec#1{\!\!\hbox{$#1$\kern-0.38em\lower0.85em\hbox{$\vec{}\,$}}\,}%
\newcommand{\bbm}{\begin{bmatrix}}
\newcommand{\ebm}{\end{bmatrix}}
\DeclareMathAlphabet{\mbf}{OT1}{ptm}{b}{n}
\newcommand{\mbs}[1]{{\boldsymbol{#1}}}

\begin{document}
%
% paper title
% can use linebreaks \\ within to get better formatting as desired
\title{Self-Calibration of the Offset Between GPS and Semantic Map Frames\\for Robust Localization}

% author names and affiliations
% use a multiple column layout for up to two different
% affiliations

\author{\IEEEauthorblockN{Wei-Kang Tseng, Angela P. Schoellig, Timothy D. Barfoot}
\IEEEauthorblockA{Institute for Aerospace Studies\\
University of Toronto\\
Toronto, Canada\\
\email{gorden.tseng@mail.utoronto.ca}, \email{schoellig@utias.utoronto.ca}, \email{tim.barfoot@utoronto.ca}}
}

% conference papers do not typically use \thanks and this command
% is locked out in conference mode. If really needed, such as for
% the acknowledgment of grants, issue a \IEEEoverridecommandlockouts
% after \documentclass

% make the title area
\maketitle

\begin{abstract}
In self-driving, standalone GPS is generally considered to have insufficient positioning accuracy to stay in lane. Instead, many turn to LIDAR localization, but this comes at the expense of building LIDAR maps that can be costly to maintain. Another possibility is to use semantic cues such as lane lines and traffic lights to achieve localization, but these are usually not continuously visible. This issue can be remedied by combining semantic cues with GPS to fill in the gaps. However, due to elapsed time between mapping and localization, the live GPS frame can be offset from the semantic map frame, requiring calibration. In this paper, we propose a robust semantic localization algorithm that self-calibrates for the offset between the live GPS and semantic map frames by exploiting common semantic cues, including traffic lights and lane markings. We formulate the problem using a modified Iterated Extended Kalman Filter, which incorporates GPS and camera images for semantic cue detection via Convolutional Neural Networks. Experimental results show that our proposed algorithm achieves decimetre-level accuracy comparable to typical LIDAR localization performance and is robust against sparse semantic features and frequent GPS dropouts.

\end{abstract}

% \begin{IEEEkeywords}
% component; formatting; style; styling;

% \end{IEEEkeywords}

% For peer review papers, you can put extra information on the cover
% page as needed:
% \ifCLASSOPTIONpeerreview
% \begin{center} \bfseries EDICS Category: 3-BBND \end{center}
% \fi
%
% For peerreview papers, this IEEEtran command inserts a page break and
% creates the second title. It will be ignored for other modes.
\IEEEpeerreviewmaketitle

\section{Introduction}
In autonomous driving applications, semantic maps have proven to be an invaluable component for most self-driving cars. They provide important prior knowledge of the surrounding environment, including the locations of drivable lanes, traffic lights, and traffic signs, as well as the traffic rules. This information is crucial for real-time behavioural planning of the vehicle under various traffic scenarios.

In order to effectively utilize semantic maps, the vehicle must be localized in the map frame down to decimetre accuracy. This proves to be challenging for the Global Positioning System (GPS), where even the best corrected version of GPS is generally considered inadequate in achieving the required accuracy consistently. Furthermore, GPS suffers from signal dropouts in situations such as inside tunnels or in dense urban environments. In light of these issues, many self-driving systems have adopted LIDAR (Light Detection and Ranging) localization methods, which require the construction of LIDAR maps prior to driving in a certain area. LIDAR localization has demonstrated great success in satisfying the stringent requirements of autonomous driving \cite{6630777,7139582,akai2017autonomous}, but this comes at the cost of building detailed geometric models of the world and keeping them up to date solely for the purpose of localization. Moreover, because the autonomous driving system, and in particular the planning component, ultimately requires the vehicle's location with respect to the semantic map, it requires the additional step of aligning the LIDAR maps with the semantic maps.

\begin{figure}[t]
    \centering
    \includegraphics[width=\columnwidth]{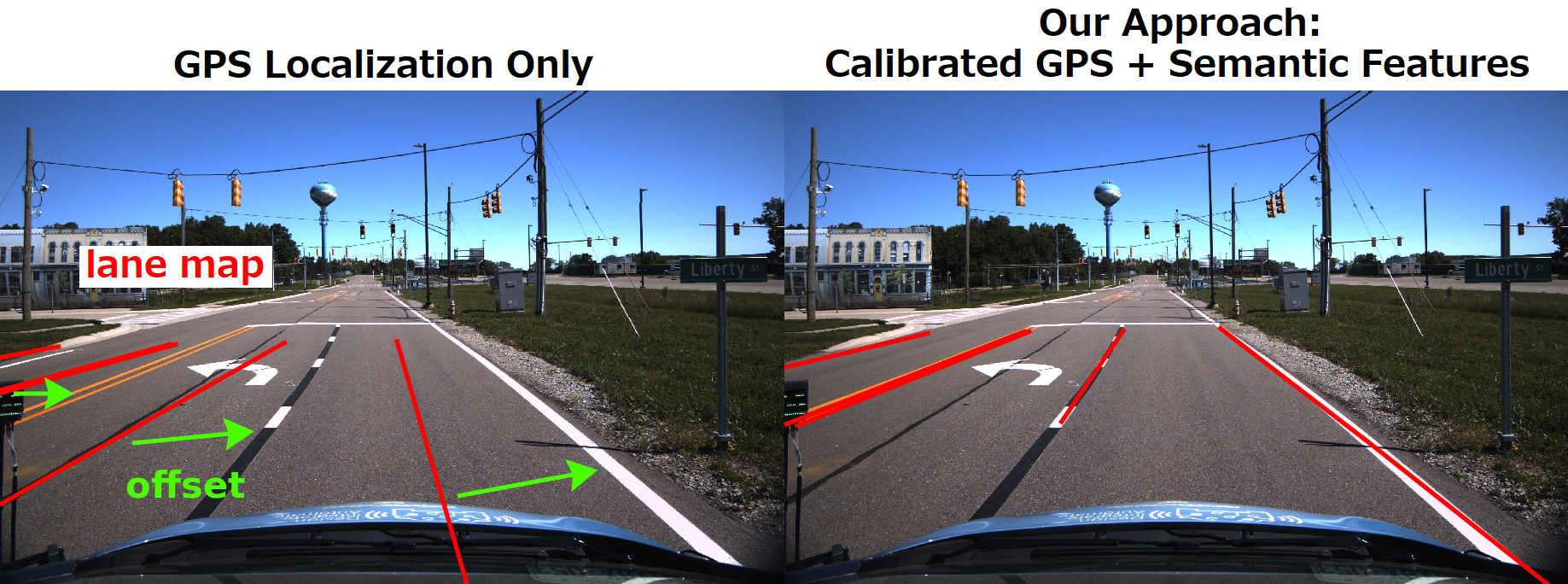}
    \caption{Vehicle localization using uncalibrated GPS (left) compared to our approach (right). The red lines are the projected lane boundaries from the semantic map. Our approach is able to self-calibrate for the GPS-to-map offset and achieve alignment between the observed lane markings and the projected lane boundaries \cite{doi:10.1002/rob.21958}.}
    \vspace{-4mm}
    \label{fig:gps_vs_semantic}
\end{figure}

An alternative to LIDAR localization is to directly take advantage of the semantic maps for localization, which the self-driving vehicle already utilizes for path planning and behavioural decision making. Through the detection of semantic cues (e.g., in the vehicle's camera images) that are also present in the semantic maps, the vehicle location can be inferred. A major downside of such an approach is that these semantic cues are fairly sparse and not always present in enough numbers to ensure reliable localization. A potential solution is to adopt a hybrid approach that combines GPS and semantic cues. However, a new problem arises: the offset between the semantic map frame and the GPS frame, in which the vehicle position is reported, must be known accurately before fusing the two sources of information. This offset is a common issue and emerges because the semantic maps are aligned to the global frame using GPS data gathered at a different time/day than when the live drive occurs. Therefore, due to different positioning of the satellites in the sky and varying atmospheric conditions \cite{laneurit:hal-00094789}, there will be a map offset requiring calibration such that the GPS frame aligns with the semantic map frame. Manual calibration is generally not practical and reliable.

As an illustrative example, aUToronto, the team that won the self-driving competition by SAE International in 2019 \cite{doi:10.1002/rob.21958}, experienced an uncalibrated GPS-to-map offset in the magnitude of a few metres, which was corrected manually just in time for the competition run, see Figure~\ref{fig:gps_vs_semantic}.

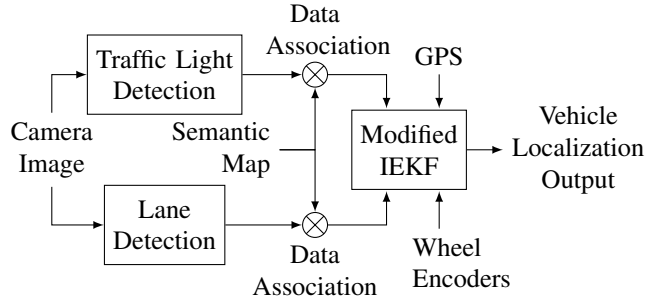
\begin{figure}[t]
    \centering
    \begingroup
        % \tikzset{every picture/.style={scale=0.1}}
        \usetikzlibrary{fit}
        \tikzset{>=latex}
\tikzset{
borderless/.style = {minimum height=0.5em, minimum width=3em,align=center},
block/.style = {draw, rectangle, minimum height=3em, minimum width=3em,align=center},
cross/.style={path picture={ 
    \draw[black](path picture bounding box.south east) -- (path picture bounding box.north west) (path picture bounding box.south west) -- (path picture bounding box.north east);
}},
da/.style= {draw, cross, circle, node distance=1cm},
outer_block/.style = {draw,dashed,inner xsep=3mm,inner ysep=2mm},
dummy/.style = {coordinate}
}

\begin{tikzpicture}[auto, node distance=3cm]
    \node [borderless] (image) {Camera\\Image};
    \node [dummy, right of=image, node distance=1.5cm] (dummy1) {};
    \node [block, above of=dummy1, node distance=1cm] (traffic_light) {Traffic Light\\Detection};
    \node [block, below of=dummy1, node distance=1cm] (lane) {Lane\\Detection};
    \node [borderless, right of=image, node distance=2.25cm, align=right] (map) {Semantic\\Map};
    \node [dummy, right of=image, node distance=3.5cm] (dummy1_1) {};
    \node [da, above of=dummy1_1, node distance=1cm] (da_light) {};
    \node [da, below of=dummy1_1, node distance=1cm] (da_lane) {};
    \node [dummy, right of=map, node distance=1.5cm] (dummy2) {};
    \node [dummy, above of=dummy2, node distance=0.55cm] (dummy3) {};
    \node [dummy, below of=dummy2, node distance=0.55cm] (dummy4) {};
    \node [dummy, right of=map, node distance=2.9cm] (dummy5) {};
    \node [dummy, right of=map, node distance=3.2cm] (dummy6) {};
    \node [borderless, above of=dummy5, node distance=1.25cm] (gps) {GPS};
    \node [borderless, below of=dummy6, node distance=1.5cm, align=left] (wheel_odom) {Wheel\\Encoders};
    \node [block, right of=map, node distance=2.5cm] (iekf) {Modified\\IEKF};
    \node [dummy, right of=dummy3, node distance=0.7cm] (iekf_dummy1) {};
    \node [dummy, right of=dummy4, node distance=0.7cm] (iekf_dummy2) {};
    \node [dummy, right of=dummy3, node distance=1.4cm] (iekf_dummy3) {};
    \node [dummy, right of=dummy4, node distance=1.4cm] (iekf_dummy4) {};
    \node [dummy, below of=iekf_dummy4, node distance=0.5cm] (wheel_odom_dummy1) {};
    \node [borderless, right of=iekf, node distance=2.25cm] (veh_loc) {Vehicle\\Localization\\Output};
    
    % arrows
    \draw [->] (image) |- (traffic_light);
    \draw [->] (image) |- (lane);
    \draw [->] (traffic_light) -- (da_light) node[above of=da_light,node distance=0.6cm,align=center]{Data\\\quad Association};
    \draw [->] (lane) -- (da_lane) node[below of=da_lane,node distance=0.6cm,align=center]{Data\\Association};
    \draw [->] (map) -| (da_light);
    \draw [->] (map) -| (da_lane);
    % \draw [-] (da_light) -| (dummy3);
    % \draw [-] (da_lane) -| (dummy4);
    % \draw [->] (dummy3) -- (iekf_dummy1);
    % \draw [->] (dummy4) -- (iekf_dummy2);
    \draw [->] (da_light) -| (iekf_dummy1);
    \draw [->] (da_lane) -| (iekf_dummy2);
    \draw [->] (gps) -- (iekf_dummy3);
    \draw [->] (wheel_odom_dummy1) -- (iekf_dummy4);
    \draw [->] (iekf) -- (veh_loc);
\end{tikzpicture}
    \endgroup
    % \vspace{-1mm}
    \caption{System architecture of our proposed localization pipeline. The camera image is passed through the lane and traffic light detectors. The data association step finds the correspondences between detection results and the semantic map projected into the image space. The results of the data association are then fused with GPS and wheel encoders in a modified IEKF to produce the final localization output.}
    \vspace{-3mm}
    \label{fig:system_overview}
\end{figure}

To address these challenges, we propose a robust localization algorithm that integrates GPS and semantic cues while performing self-calibration of the offset between the GPS and semantic map frames. By folding the offset into our state estimation, we can properly fuse the two sources of information while benefitting from both. For this work, we assume detection of semantic cues using a front-facing monocular camera, and formulate the localization problem as a modified Iterated Extended Kalman Filter (IEKF), which improves upon the linearization of EKF. The system architecture is summarized in Figure~\ref{fig:system_overview}.

The proposed approach has minimal computational impact because GPS is low-cost to process, and common semantic cues such as lane markings and traffic lights are already tracked for the purpose of vehicle behavioural planning, so the added cost of using them is also low. The result is an accurate and robust self-driving localization pipeline that uses GPS to fill in the gaps between sparse semantic observations, avoids the need for expensive maps specifically for localization, and relies on features in the environment that are actively maintained and designed to be highly visible. Experimental results in an urban environment using the Carla simulator \cite{Dosovitskiy17} as well as on a real-world dataset collected by aUToronto during the SAE AutoDrive competition show that we are able to achieve 3~cm lateral and 5~cm longitudinal accuracy on average, and also maintain similar performance with frequent GPS dropouts.

The paper is organized as follows. Section~\ref{sec:related_work} summarizes the related work on vehicle localization. Section~\ref{sec:semantic_preprocessing} describes the preprocessing required for the semantic cues. Section~\ref{sec:vehicle_localization} presents the mathematical formulation of the localization algorithm. Section~\ref{sec:experiments} provides the experimental results. Finally, Section~\ref{sec:conclusion} concludes the paper.

\section{RELATED WORK}
\label{sec:related_work}
\subsubsection{LIDAR Localization}
One of the most popular localization approaches in self driving is LIDAR localization \cite{4341827,hata2015feature,egger2018posemap,le2020in2laama}. By constructing a database of the detailed geometry of the environment in advance, localization can be achieved using a point cloud registration algorithm, which matches the LIDAR scans against the database at test time. Because the localization performance greatly depends on the accuracy of the database in capturing the ever-changing appearance of the world, the database needs to be frequently updated. In response, many have developed algorithms that extract features that are more invariant to environmental changes in the LIDAR data \cite{yoneda2014lidar,im2016vertical,liu2019precise}. More recently, \cite{lu2019l3} proposed a novel learning-based approach that directly takes LIDAR point clouds as inputs and learns descriptors for matching in various driving scenarios. While these methods help mitigate the impact of outdated LIDAR database, they do not fundamentally address the issue of needing to maintain a separate database solely for localization.

\subsubsection{Semantic Localization}
Semantic localization exploits various common roadside semantic cues present in the semantic maps to achieve vehicle localization. In contrast to LIDAR localization, this method conveniently makes use of the same semantic maps already required by the autonomous vehicle for planning purposes. Therefore, no maintenance of a separate database of the environment is required. Among the various types of semantic cues, lane markings are most commonly utilized because they are abundant and provide important clues that keep the vehicle in the correct lane \cite{6856528,gruyer2016accurate,1505109,vivacqua2017low,7313499}. However, since lane markings tend to run parallel to the vehicle heading, the longitudinal localization accuracy is usually worse than lateral accuracy. Besides lane markings, other types of semantic cues have been exploited as well, including stop lines \cite{6629509,6387308}, other road markings \cite{7160754,6629627,7547970}, traffic lights \cite{6166893,8412763}, and traffic signs \cite{7313530,7225751,7995692,choi2018fast}. A common issue that all types of semantic cues suffer from is sparsity. In response, approaches that combine multiple types of semantic cues have been proposed, most of which include lane markings in combination with traffic lights or traffic signs \cite{ma2019exploiting,8868068,5625240}. Many of the semantic localization papers referenced in this section have incorporated GPS into their localization pipelines, but none of them addressed a possible offset between GPS and semantic map frames due to reasons discussed above, presumably because the GPS offset has been manually corrected prior to experiments. However, as experienced by aUToronto, manual GPS calibration is often unreliable, and can lead to localization failures \cite{doi:10.1002/rob.21958}.

\subsubsection{GPS Calibration with Semantic Cues}
In this work, the GPS measurements are regarded as reporting the vehicle positions with respect to a GPS frame, which is at an offset from the semantic map frame. Alternatively, we can treat the GPS as if it directly reports the vehicle position in the semantic map frame, but with a systematic bias. Some prior works took this fact into account when developing their semantic localization pipelines. For instance, \cite{10.3390/s150820779} simply modelled the GPS errors as a random constant since the change in the GPS bias is small. A more sophisticated model utilizing autoregressive process such as a random walk was shown by \cite{6696383} to achieve superior performance compared to the random constant model, and was similarly adopted by \cite{6629538} and \cite{7353808}. All of these approaches only adopted road markings as the semantic cues. While our approach is similar in spirit to these papers, there are also notable differences, including the addition of traffic lights as part of the semantic cues, and their detections using Convolutional Neural Networks (CNNs). Furthermore, we formulate the localization problem in 3D, process the semantic cue detection results directly in the image space rather than in bird's-eye view, and ensure robust localization against frequent GPS dropouts.

\section{SEMANTIC CUE PREPROCESSING}
\label{sec:semantic_preprocessing}
\subsubsection{Semantic Map}
Our localization algorithm utilizes a lightweight HD semantic map that consists of a lane graph and traffic light locations. A lane graph is a set of polylines that defines all the lane boundaries of the road network. It corresponds to visually distinctive lane markings, which can be easily identified in a camera image. The traffic lights are treated as point landmarks with the coordinates of their centres recorded in the semantic map. In this work, we assume the semantic map has been provided.

\subsubsection{Traffic Light Detection}
Traffic lights appear regularly at road intersections and provide useful information for longitudinal localization. A traffic light CNN detector \cite{redmon2018yolov3} outputs bounding boxes that locate traffic lights in the camera images. The centre of each bounding box is then obtained as the observed point landmark of a traffic light. To filter out false detections, only bounding boxes with a high confidence level are included for localization.

Before the traffic light detections can be made useful for localization, a data association scheme must first be devised to correctly associate the detections in the image with corresponding traffic lights in the semantic map. This is achieved by first projecting the locations of all nearby traffic lights in the semantic map to the image space using the estimated vehicle position. We then apply Iterative Closest Point followed by nearest neighbour to obtain the desired associations. Detections that have no nearby associations within a certain distance threshold are identified as outliers and discarded. Figure~\ref{fig:data_association} illustrates the results of the traffic light data association process.

\subsubsection{Lane Marking Detection}
\label{sec:lane_detection}
Lane markings are one of the most common type of semantic cues that primarily help with lateral localization. A lane marking CNN detector \cite{Takikawa2019GatedSCNNGS} produces a mask of the camera image that classifies each pixel as lane marking with a probability. A binary mask is then obtained by applying a probability threshold. Only the bottom portion of the mask, where the lane markings are closer to the vehicle and can be clearly identified, is retained. The resulting image coordinates of the pixels classified as lane markings are then evenly subsampled to reduce computational burden.

\begin{figure}[t]
    \centering
    \includegraphics[trim=0 0 0 150,clip,width=\columnwidth]{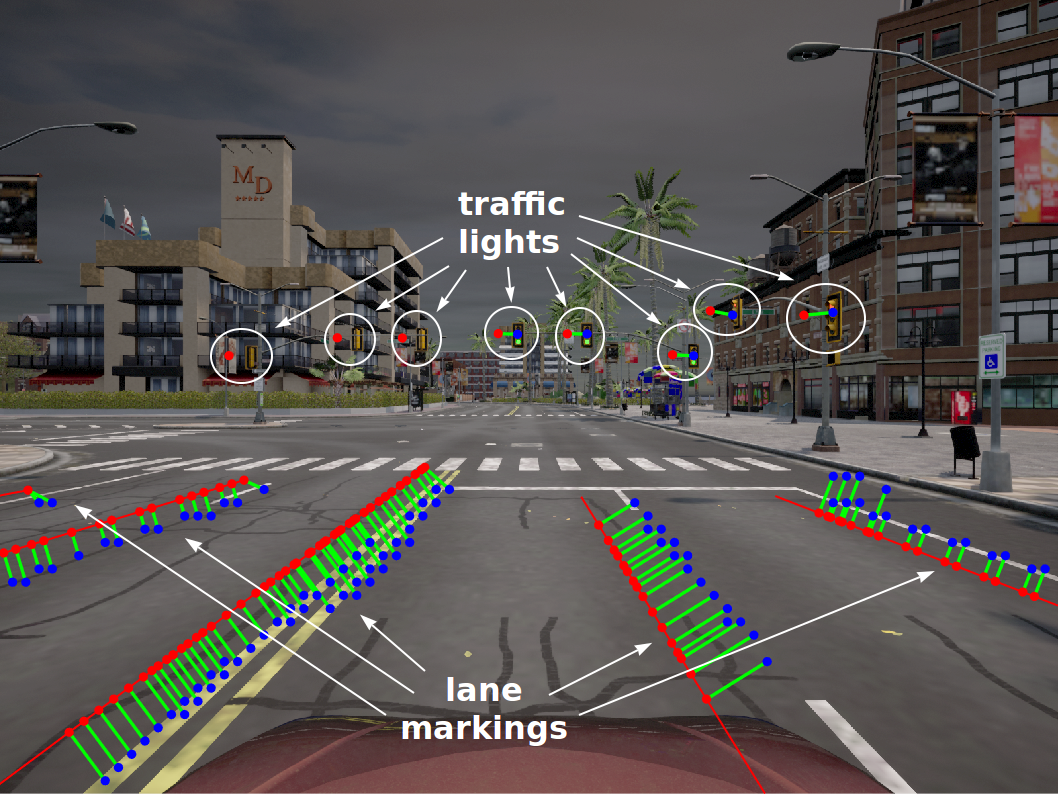}
    \vspace{-5mm}
    \caption{Data association process of traffic lights and lane markings. The red lines and points are known positions of semantic cues projected into the image using the estimated vehicle position; the blue points are semantic cue detections; and the green lines indicate the matching results. A few incorrect matches of outlier lane pixels can be observed on the right due to their proximity to a projected lane line.}
    \vspace{-4mm}
    \label{fig:data_association}
\end{figure}

The data association process for lane markings also begins by projecting all nearby lane lines from the semantic map to the image space. Next, the subsampled lane pixels are each matched to their closest projected lines. Outlier lane pixels without nearby matches are discarded. Figure~\ref{fig:data_association} demonstrates the lane marking matching results. Given all the lane pixels matched to each projected line, a straight line is fitted using least squares in image space. Finally, data association is obtained between the pairs of fitted lines from lane marking detection and projected lines from the semantic map.% To avoid lane misassociations for those lanes farther away from the ego-lane, we also discard all the fitted lines that do not intersect with the bottom of the image.

\section{VEHICLE LOCALIZATION}
\label{sec:vehicle_localization}

\subsection{Problem Setup}
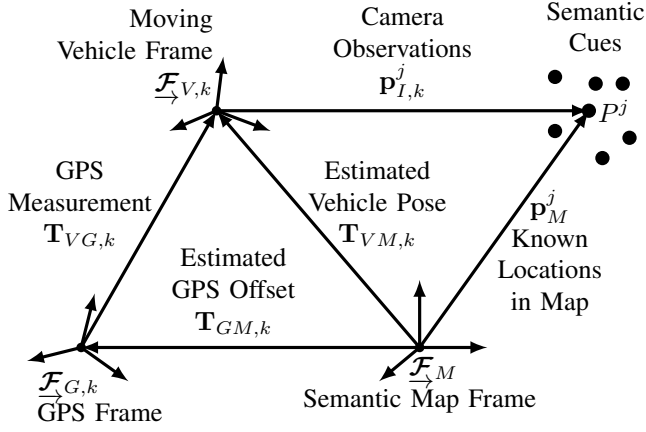
\begin{figure}[t]
    \centering
    \begingroup
        \tikzset{every picture/.style={scale=0.9}}
        % - Custom Math Commands ---------------------------------------
\newcommand{\refframe}{\underrightarrow{\bm{\mathcal{F}}}}

\tikzset{>=latex}
\begin{tikzpicture}
    % Map reference frame
    \newcommand\MapXoffset{5};
    \newcommand\MapYoffset{0};
    \draw[very thick,->] (0+\MapXoffset,0+\MapYoffset) -- (1+\MapXoffset,0+\MapYoffset);
    \draw[very thick,->] (0+\MapXoffset,0+\MapYoffset) -- (0+\MapXoffset,1+\MapYoffset);
    \draw[very thick,->] (0+\MapXoffset,0+\MapYoffset) -- (-0.6+\MapXoffset,-0.5+\MapYoffset);
    \fill (0+\MapXoffset,0+\MapYoffset) circle (2pt) node[align=center, anchor=north] {\quad $\refframe_M$\\Semantic Map Frame};
    
    % Vehicle reference frame
    \newcommand\VehicleXoffset{2};
    \newcommand\VehicleYoffset{3.5};
    \draw[very thick,->] (0+\VehicleXoffset,0+\VehicleYoffset) -- (0.8+\VehicleXoffset,-0.3+\VehicleYoffset);
    \draw[very thick,->] (0+\VehicleXoffset,0+\VehicleYoffset) -- (0.1+\VehicleXoffset,0.8+\VehicleYoffset) node[align=right, anchor=east] {Moving\\Vehicle Frame\\$\refframe_{V,k}$};
    \draw[very thick,->] (0+\VehicleXoffset,0+\VehicleYoffset) -- (-0.7+\VehicleXoffset,-0.3+\VehicleYoffset);
    \fill (0+\VehicleXoffset,0+\VehicleYoffset) circle (2pt);
    
    % GPS reference frame
    \draw[very thick,->] (0,0) -- (0.2,0.8);
    \draw[very thick,->] (0,0) -- (0.7,-0.5);
    \draw[very thick,->] (0,0) -- (-0.8,-0.2) node[align=left, anchor=north west] {$\refframe_{G,k}$\\GPS Frame};
    \fill (0,0) circle (2pt);
    
    % Landmarks
    \newcommand\LandmarkXoffset{7.5};
    \newcommand\LandmarkYoffset{3.5};
    \fill (0+\LandmarkXoffset,0+\LandmarkYoffset) circle (3pt) node[right] {$P^j$};
    \fill (0.6+\LandmarkXoffset,-0.4+\LandmarkYoffset) circle (3pt);
    \fill (0.5+\LandmarkXoffset,0.4+\LandmarkYoffset) circle (3pt);
    \fill (-0.5+\LandmarkXoffset,0.5+\LandmarkYoffset) circle (3pt);
    \fill (0.1+\LandmarkXoffset,0.4+\LandmarkYoffset) circle (3pt) node[align=center, above = 0.3cm] {Semantic\\Cues};
    \fill (0.2+\LandmarkXoffset,-0.7+\LandmarkYoffset) circle (3pt);
    \fill (-0.5+\LandmarkXoffset,-0.3+\LandmarkYoffset) circle (3pt);
    
    % Arrows between reference frames and landmark
    \draw[very thick,->] (0,0) -- (0+\VehicleXoffset,0+\VehicleYoffset) node[align=center, pos=0.6, left] {GPS\\Measurement\\$\mathbf{T}_{VG,k}$};
    \draw[very thick,->] (0+\MapXoffset,0+\MapYoffset) -- (0,0) node[align=center, pos=0.55, above] {Estimated\\GPS Offset\\$\mathbf{T}_{GM,k}$};
    \draw[very thick,->] (0+\MapXoffset,0+\MapYoffset) -- (0+\VehicleXoffset,0+\VehicleYoffset) node[align=center, pos=0.6,right] {Estimated\\Vehicle Pose\\$\mathbf{T}_{VM,k}$};
    \draw[very thick,->] (0+\MapXoffset,0+\MapYoffset) -- (0\LandmarkXoffset,0+\LandmarkYoffset) node[align=center, pos=0.4, right] {$\mathbf{p}_M^j$\\Known\\Locations\\in Map};
    \draw[very thick,->] (0+\VehicleXoffset,0+\VehicleYoffset) -- (0\LandmarkXoffset,0+\LandmarkYoffset) node[align=center, midway, above] {Camera\\Observations\\$\mathbf{p}_{I,k}^j$};
\end{tikzpicture}
    \endgroup
    \vspace{-2mm}
    \caption{Definition of reference frames for the localization problem with semantic cues and offset between GPS and semantic map frames.}
    \vspace{-5mm}
    \label{fig:problem_setup}
\end{figure}

We formulate the semantic localization problem with GPS offset by first discretizing the time denoted by subscript $k$. There are three reference frames. $\bcfa_M$ is the semantic map frame, $\bcfa_{V,k}$ is attached to a moving vehicle, and $\bcfa_{G,k}$ is the GPS frame, which is at an offset from $\bcfa_M$. We then have three corresponding transformation matrices between the frames. $\mathbf{T}_{VG,k} \in SE(3)$ is the GPS measurement of the pose of vehicle, $\mathbf{T}_{GM,k} \in SE(3)$ is the GPS-to-map offset, which needs to be estimated for self-calibration, and $\mathbf{T}_{VM,k} \in SE(3)$ is the pose of vehicle with respect to the semantic map, which we ultimately desire. Figure~\ref{fig:problem_setup} illustrates the described problem setup. At time step $k$, the $j$-th semantic cue, $P^j$, detected by the onboard camera has the pixel coordinates, $\mathbf{p}_{I,k}^j$, as well as its known location in the map frame, $\mathbf{p}_M^j \in \Real^{3}$, obtained from the semantic map. Using $\mathbf{T}_{VM,k}$, we can transform and project the known location, $\mathbf{p}_M^j$, to the image space and obtain the reprojection error for localization and GPS-to-map offset calibration.

We adopt the mathematical notations from \cite{Barfoot:2017}. Notably, $\wedge$ is an operator associated with the Lie algebra for $SE(3)$:
\begin{equation}
    \mbs{\xi}^{\wedge}
    =
    \bbm
        \mbs{\rho} \\
        \mbs{\phi}
    \ebm
    ^{\wedge}
    :=
    \bbm
        \mbs{\phi}^{\wedge} & \mbs{\rho} \\
        \mbs{0}^T & 0
    \ebm
    \in
    \Real^{4 \times 4}
    ,\;\;\;\;
    \mbs{\rho}, \mbs{\phi}
    \in
    \Real^3,
\end{equation}
and also for $SO(3)$:
\begin{equation}
    \mbs{\phi}^{\wedge}
    =
    \bbm
        \phi_1 \\
        \phi_2 \\
        \phi_3
    \ebm
    ^{\wedge}
    :=
    \bbm
        0 & -\phi_3 & \phi_2 \\
        \phi_3 & 0 & -\phi_1 \\
        -\phi_2 & \phi_1 & 0
    \ebm.
\end{equation}

\subsection{Process and Observation Models}

\subsubsection{Vehicle Dynamics Process Model}
We adopt the white-noise-on-acceleration model \cite{7353368}, which also estimates the vehicle velocity $\mbs{\varpi}_k \in \Real^{6}$ in the vehicle frame $\bcfa_{V,k}$:
\begin{align}
    \mathbf{T}_{VM,k} &=
    \exp(\mathbf{w}_{VM}^{\wedge})
    \exp(\Delta t_k \mbs{\varpi}_{k-1}^{\wedge})
    \mathbf{T}_{VM,k-1},
    \label{eq:vehicle_pose_process}
    \\
    \mbs{\varpi}_k &= \mbs{\varpi}_{k-1} + \mathbf{w}_{\varpi},
    \label{eq:vehicle_velocity_process}
\end{align}
where $\Delta t_k = t_k - t_{k-1}$ is the time interval, and $\mathbf{w}_{VM}$, $\mathbf{w}_{\varpi} \in \Real^{6}$ are zero mean Gaussian process noises for vehicle pose and velocity, respectively. As formulated in \cite{7353368}, $\mathbf{w}_{VM}$ and $\mathbf{w}_{\varpi}$ are correlated with the joint distribution
\begin{equation}
    \bbm
        \mathbf{w}_{VM} \\
        \mathbf{w}_{\varpi}
    \ebm
    \isGaussianBigg{
        \mbs{0}_{12 \times 1}
    }{
        \underbrace{
            \bbm
                \frac{1}{3}\Delta t_k^3\mathbf{Q}_C & \frac{1}{2}\Delta t_k^2\mathbf{Q}_C \\
                \frac{1}{2}\Delta t_k^2\mathbf{Q}_C & \Delta t_k\mathbf{Q}_C \\
            \ebm
        }_{
            \mathbf{Q}_{VM}
        }
    },
\end{equation}
where the tunable parameter $\mathbf{Q}_C \in \Real^{6 \times 6}$ is a diagonal matrix with non-zero values in its first and last diagonal entries corresponding to the vehicle's translational and rotational accelerations in the vehicle frame, which are along the $x$-axis (tangential to its motion) and about the $z$-axis (normal to the ground plane), respectively.

\subsubsection{GPS Offset Process Model}
The GPS-to-map offset, which very gradually varies over time, is modelled as a random walk. This is a convenient way to handle the estimation of such a time-dependent unknown parameter:
\begin{equation}
    \mathbf{T}_{GM,k} = \exp(\mathbf{w}_{GM}^{\wedge})\mathbf{T}_{GM,k-1},
    \label{eq:gps_offset_process}
\end{equation}
where $\mathbf{w}_{GM} \isGaussian{\mbs{0}_{6 \times 1}}{\mathbf{Q}_{GM}}$ is the process noise.

\subsubsection{GPS Observation Model}
In this work, a GPS measurement refers to a preprocessed quantity that is a three-dimensional transformation matrix $\mathbf{T}_{VG,k}$ with three degrees of freedom each in position and orientation. This is the output of commercial GPS-based localization solutions such as Applanix POS LV, which integrates GPS and IMU information. The observation model of GPS measurement, $\mathbf{T}_{VG,k}$, is
\begin{equation}
    \mathbf{T}_{VG,k} = \exp(\mathbf{n}_{VG}^\wedge)\mathbf{T}_{VM,k}\mathbf{T}_{GM,k}^{-1},
    \label{eq:joint_obs_gps}
\end{equation}
where the measurement noise is $\mathbf{n}_{VG} \isGaussian{\mathbf{0}_{6 \times 1}}{\mathbf{R}_{VG}}$.

\subsubsection{Traffic Light Observation Model}
For the $j$-th traffic light pixel measurement, the observation model is simply
\begin{equation}
    \mathbf{p}_{I,k}^j=\mathbf{g}(\mathbf{p}_M^j,\mathbf{T}_{VM,k})+\mathbf{n}_{\text{light}},
    \label{eq:joint_obs_light}
\end{equation}
where $\mathbf{g}(\cdot)$ projects the known traffic light location $\mathbf{p}_M^j$ from semantic map to image space of the onboard camera given vehicle pose estimation, $\mathbf{T}_{VM,k}$. The pixel measurement noise $\mathbf{n}_{\text{light}} \isGaussian{\mathbf{0}_{2 \times 1}}{\mathbf{R}_{\text{light}}}$ is assumed to be Gaussian.

\subsubsection{Lane Marking Observation Model}
Using the data association process described in Section~\ref{sec:lane_detection}, we obtain lane marking observations as straight lines in the image space, which can each be represented by two distinct points on the line. For the $j$-th observed line, the points are selected by choosing two different vertical pixel coordinates $\mathbf{y}_I^j = \bbm y_1^j & y_2^j\ebm^T$. The observation model for the corresponding horizontal pixel coordinates is
\begin{equation}
    \mathbf{x}_{I,k}^j
    =
    \bbm
        x_{1,k}^j & x_{2,k}^j
    \ebm
    ^T
    =\mathbf{f}(\mathbf{g}(\boldsymbol{\ell}_M^j,\mathbf{T}_{VM,k}),\mathbf{y}_I^j)+\mathbf{n}_{\text{lane}},
    \label{eq:joint_obs_lane}
\end{equation}
where $\boldsymbol{\ell}_M^j$ is the known lane line from the semantic map, and $\mathbf{f}(\cdot)$ produces the horizontal pixel coordinates given $\mathbf{y}_I^j$. The associated measurement noise is $\mathbf{n}_{\text{lane}} \isGaussian{\mathbf{0}_{2 \times 1}}{\mathbf{R}_{\text{lane}}}$.

\subsubsection{Wheel Encoders Observation Model}
The onboard wheel encoders provide measurements on the vehicle's longitudinal velocity, $v_k$, and angular velocity, $\omega_k$, in yaw. Wheel encoders are included to improve robustness against GPS dropouts. The observation model is
\begin{equation}
    \mbs{\varpi}_{\text{wheel},k}
    =
    \bbm
        v_k & \omega_k
    \ebm
    ^T
    =
    \mathbf{h}_{\varpi}(\mbs{\varpi}_k) + \mathbf{n}_{\varpi},
    \label{eq:joint_obs_encoder}
\end{equation}
where $\mathbf{h}_{\varpi}(\cdot)$ extracts the corresponding vehicle velocities from $\mbs{\varpi}_k$, and the noise term is $\mathbf{n}_{\varpi} \isGaussian{\mathbf{0}_{2 \times 1}}{\mathbf{R}_{\varpi}}$.

\subsubsection{Pseudo-Measurement Observation Model}
Some pseudo-measurements are introduced by leveraging the fact that the vehicle always stays on the ground, which is assumed to be the $xy$-plane of the map frame. Therefore, its elevation, roll, and pitch are all soft-constrained to zero with respect to the map. This effectively reduces the localization problem down to a 2D space while maintaining the problem formulation in 3D. Additionally, we assume that the rear wheels do not slip sideways, thus the lateral vehicle velocity is also near zero. The observation models of the pseudo-measurements are straightforward:
\begin{equation}
    u_{\text{pseudo}} = h_u(\mathbf{T}_{VM,k},\mbs{\varpi}_k) + n_u,
    \label{eq:joint_obs_pseudo}
\end{equation}
where $u$ is elevation, roll, or pitch of the vehicle with respect to the map frame, or lateral velocity in the vehicle frame. $h_u(\cdot)$ extracts the corresponding quantity from the current vehicle pose and velocity estimations. The corresponding pseudo-measurement noise is $n_u \isGaussian{0}{r_u}$, with small $r_u$ to effectively constraint each quantity.

\subsection{Modified IEKF Formulation}

\subsubsection{Prediction Step}
The prediction step follows the standard IEKF formulation by jointly estimating the vehicle pose, velocity, and GPS offset using the process models \eqref{eq:vehicle_pose_process}, \eqref{eq:vehicle_velocity_process}, and \eqref{eq:gps_offset_process}. We linearize them to obtain the predicted means
\begin{align}
    \check{\mathbf{T}}_{VM,k} &= 
    \exp(\Delta t_k \hat{\mbs{\varpi}}_{k-1}^{\wedge})\hat{\mathbf{T}}_{VM,k-1},
    \label{eq:vehicle_pose_predicted_mean} \\
    \check{\mbs{\varpi}}_{k} &= \hat{\mbs{\varpi}}_{k-1},
    \label{eq:vehicle_velocity_predicted_mean} \\
    \check{\mathbf{T}}_{GM,k} &= \hat{\mathbf{T}}_{GM,k-1},
    \label{eq:gps_offset_predicted_mean}
\end{align}
as well as the predicted joint covariance matrix
\begin{equation}
    \check{\mathbf{P}}_k
    =
    \mathbf{F}_{k-1}
    \hat{\mathbf{P}}_{k-1}
    \tp{\mathbf{F}_{k-1}}
    +
    \bbm
        \mathbf{Q}_{VM} & \mathbf{0}_{12 \times 6} \\
        \mathbf{0}_{6 \times 12} & \mathbf{Q}_{GM}
    \ebm,
    \label{eq:joint_predicted_covar}
\end{equation}
where $\mathbf{F}_{k}$ is the combined Jacobian matrix of the linearized process models \eqref{eq:vehicle_pose_process}, \eqref{eq:vehicle_velocity_process}, and \eqref{eq:gps_offset_process} at time step $k$.

\subsubsection{Correction Step}
The iterative correction step of IEKF is modified by replacing it with a batch optimization formulation with time window size of one (the current time step) \cite{7353368}. The cost function to optimize is $J = J_v + J_y$, where
\begin{align}
    J_v &= \frac{1}{2}\tp{\mathbf{e}_{v,k}} \check{\mathbf{P}}_k^{-1} \mathbf{e}_{v,k}, \\
    J_y &= \sum_{i}\frac{1}{2}\tp{\mathbf{e}_{y,k}^i} {\mathbf{R}^i}^{-1} \mathbf{e}_{y,k}^i,
    \label{eq:corr_step_meas_cost}
\end{align}
are the prior and measurement cost terms, respectively. The prior errors, $\mathbf{e}_{v,k}$, are computed using the predicted means, $\check{\mathbf{T}}_{VM,k}$, $\check{\mbs{\varpi}}_{k}$, and $\check{\mathbf{T}}_{GM,k}$, from the prediction step \eqref{eq:vehicle_pose_predicted_mean}, \eqref{eq:vehicle_velocity_predicted_mean}, and \eqref{eq:gps_offset_predicted_mean}. This encourages a consistent trajectory that respects the vehicle dynamics between the estimated poses. The overall measurement cost term $J_y$ is a sum of the cost terms derived from the sensor measurements, including GPS \eqref{eq:joint_obs_gps}, semantic cues \eqref{eq:joint_obs_light}\eqref{eq:joint_obs_lane}, wheel encoders \eqref{eq:joint_obs_encoder}, and pseudo-measurements \eqref{eq:joint_obs_pseudo}. For the $i$-th measurement, $\mathbf{e}_{y,k}^i$ is the measurement error and $\mathbf{R}^i$ is the associated observation covariance matrix: $\mathbf{R}_{VG}$,  $\mathbf{R}_{\text{light}}$, $\mathbf{R}_{\text{lane}}$, $\mathbf{R}_{\varpi}$, or $r_u$.

In order to minimize the impact of bad data association of semantic cues, a Cauchy M-estimator is deployed \cite{hampel1986robust}. For each semantic cue measurement cost term in \eqref{eq:corr_step_meas_cost}, ${\mathbf{R}^i}^{-1}$ is replaced by ${\mathbf{Y}_{k}^i}^{-1} = (1+\tp{\mathbf{e}_{y,k}^i} {\mathbf{R}^i}^{-1} \mathbf{e}_{y,k}^i)^{-1} {\mathbf{R}^i}^{-1}$.
% \begin{equation}
%     {\mathbf{Y}_{k}^i}^{-1} = (1+\tp{\mathbf{e}_{y,k}^i} {\mathbf{R}^i}^{-1} \mathbf{e}_{y,k}^i)^{-1} {\mathbf{R}^i}^{-1}.
% \end{equation}
Given a reasonably well initialized vehicle position, this scheme effectively prevents localization failures by scaling down the importance of outliers, which produce large measurement errors, via the associated ${\mathbf{Y}_{k}^i}^{-1}$.

The cost function, $J$, is optimized using the Gauss-Newton method. In each iteration, we first obtain the combined Jacobian matrices, $\mathbf{E}_{k}$ and $\mathbf{G}_{k}$, after linearizing the error terms, $\mathbf{e}_{v,k}$ and $\mathbf{e}_{y,k}^i$, about the operating point, $\mathbf{x_{\text{op}}} = \{ \hat{\mathbf{T}}_{VM,\text{op},k}$, $\hat{\mbs{\varpi}}_{\text{op},k}$, $\hat{\mathbf{T}}_{GM,\text{op},k} \}$. We then substitute the linearized error terms into the cost function and set its derivative with respect to the perturbation term, $\delta\mathbf{x} = \bbm \delta\mbs{\xi}_{VM,k}^T & \delta\mbs{\varpi}_k^T & \delta\mbs{\xi}_{GM,k}^T \ebm^T \in \Real^{18}$, to zero. This produces the update equation
\begin{align}
    (
    \overbrace{
        \mathbf{E}_{k}^{T} \check{\mathbf{P}}_k^{-1} \mathbf{E}_{k}
    }^{
        \mathbf{A}_{\text{pri}}
    }
    +
    \overbrace{
        \mathbf{G}_{k}^T \mathbf{R}_k^{-1} \mathbf{G}_{k}
    }^{
        \mathbf{A}_{\text{meas}}
    }
    )
    \delta\mathbf{x}^*
    &=
    \nonumber
    \\
    \mathbf{E}_{k}^{T} \check{\mathbf{P}}_k^{-1} \mathbf{e}_{v,k}
    &+
    \mathbf{G}_{k}^T \mathbf{R}_k^{-1} \mathbf{e}_{y,k}
    ,
\end{align}
where $\delta\mathbf{x}^*$ is the optimal perturbation, and $\mathbf{R}_k$ is a block diagonal matrix that combines all $\mathbf{R}^i$ and $\mathbf{Y}_k^i$. Solving for $\delta\mathbf{x}^*$, we update the operating point as follows:
\begin{align}
    \hat{\mathbf{T}}_{VM,\text{op},k} &\leftarrow \exp((\delta\mbs{\xi}_{VM,k}^*)^\wedge)\hat{\mathbf{T}}_{VM,\text{op},k}
    ,\\
    \hat{\mbs{\varpi}}_{\text{op},k} &\leftarrow \hat{\mbs{\varpi}}_{\text{op},k} + \delta\mbs{\varpi}_k^*
    ,\\
    \hat{\mathbf{T}}_{GM,\text{op},k} &\leftarrow \exp((\delta\mbs{\xi}_{GM,k}^*)^\wedge)\hat{\mathbf{T}}_{GM,\text{op},k}
    .
\end{align}
Finally, the results of the correction step, $\hat{\mathbf{T}}_{VM,k}$, $\hat{\mbs{\varpi}}_{k}$, and $\hat{\mathbf{T}}_{GM,k}$, are output after convergence. The corresponding covariance is computed as $\hat{\mathbf{P}}_k = (\mathbf{A}_{\text{pri}} + \mathbf{A}_{\text{meas}})^{-1}$ from the last iteration.

\section{EXPERIMENTS}
\label{sec:experiments}
Quantitative simulations were carried out to verify the method using the Carla simulator \cite{Dosovitskiy17}.  We also gathered anecdotal results from a real-world dataset to validate the feasibility of our approach in reality.

\subsection{Parameter Tuning}
Our localization pipeline involves numerous parameters, including the outlier distance threshold in data association, and matrices related to the state estimator. In the process models, $\mathbf{Q}_C$ is associated with the state covariance of vehicle pose and velocity, and $\mathbf{Q}_{GM}$ affects the magnitude of the random walk of GPS frame. The observation noise parameters consist of $\mathbf{R}_{VG}$, $\mathbf{R}_{\text{light}}$, $\mathbf{R}_{\text{lane}}$, $\mathbf{R}_{\varpi}$, and $r_u$ that are associated with GPS, traffic light, lane markings, wheel encoders, and pseudo-measurements, respectively. These parameters are manually tuned by initializing them with reasonable values, followed by adjustments to achieve optimal performance evaluated on a validation dataset generated from Carla.

\subsection{Carla Simulation}
\begin{figure}[t]
    \centering
    \includegraphics[trim=0 5 0 5,clip,width=0.96\columnwidth]{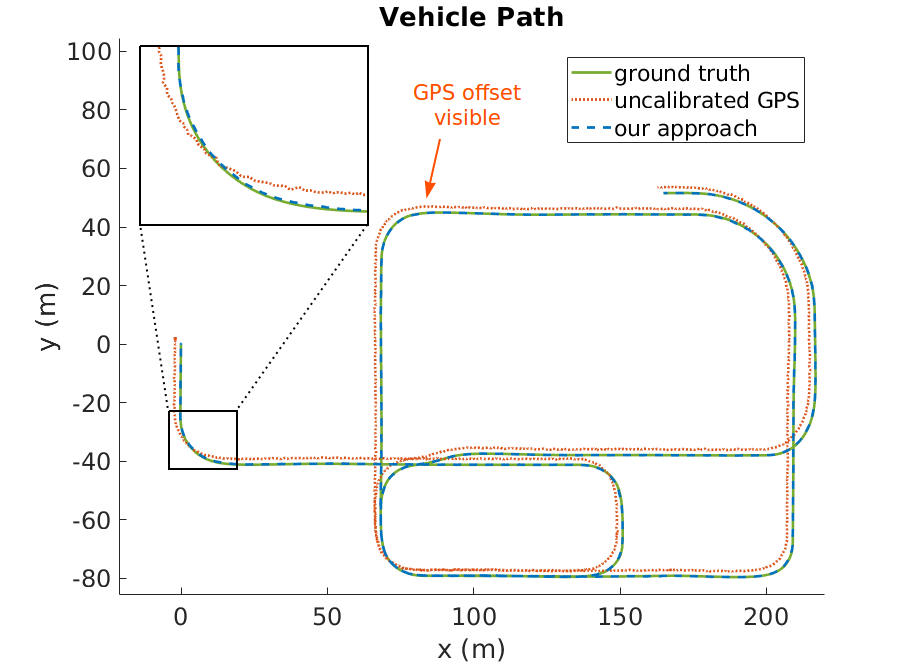}
    \vspace{-2mm}
    \caption{Vehicle path of Carla simulation with total length of roughly 1~km. The ground truth path is compared with results from uncalibrated GPS and our proposed approach. One of the turns is zoomed in to show that the estimated path using our approach very closely overlap with the ground truth path, while the uncalibrated GPS path significantly diverges from it.}
    \vspace{-4mm}
    \label{fig:vehicle_path}
\end{figure}

Our localization algorithm was first tested using Carla, an autonomous driving simulator \cite{Dosovitskiy17}. In particular, the experiments were conducted using the map ``Town10HD'', which offers a photorealistic urban driving environment and a perfect semantic map. The benefit of using a simulator is the availability of ground truth, which simplifies the analysis of localization results. Due to Carla not supporting an offset in GPS measurements, we instead manually injected one with 2~m in both longitude and latitude. The time-dependency of the offset is ignored since it is negligible in the duration of a simulation run. With this setup, the simulation data was obtained from roughly 1~km of driving. The vehicle path comparing ground truth with our method and uncalibrated GPS is shown in Figure~\ref{fig:vehicle_path}.

\subsubsection{Localization Results}
The longitudinal, lateral, and heading localization errors computed using ground truth are summarized in Table~\ref{table:carla_results}. Our proposed method achieves highly accurate results with a median longitudinal error of 0.053~m, a median lateral error of 0.031~m, and a median heading error of 0.004~radians. When shown as a histogram in Figure~\ref{fig:carla_hist}, we observe that the longitudinal errors have a larger spread than lateral errors, and the vehicle heading always remains very accurate. This is in line with our expectations since lane markings, the most abundant type of semantic cues, only provide lateral and heading corrections. In contrast, longitudinal corrections offered by traffic lights are only available around road intersections.

\subsubsection{GPS Offset Estimation}
Being the key motivation for developing the proposed localization algorithm, achieving accurate estimation of the GPS-to-map offset is crucial. The blue line in Figure~\ref{fig:carla_gps_offset} shows the GPS offset estimation error. Starting from a poor initial guess, our localization algorithm successfully refines the estimates and drops the error down to just a few centimetres, with no manual calibration required.

\subsubsection{GPS Dropouts}
\begin{figure}[t]
    \centering
    \includegraphics[trim=0 40 0 35,clip,width=\columnwidth]{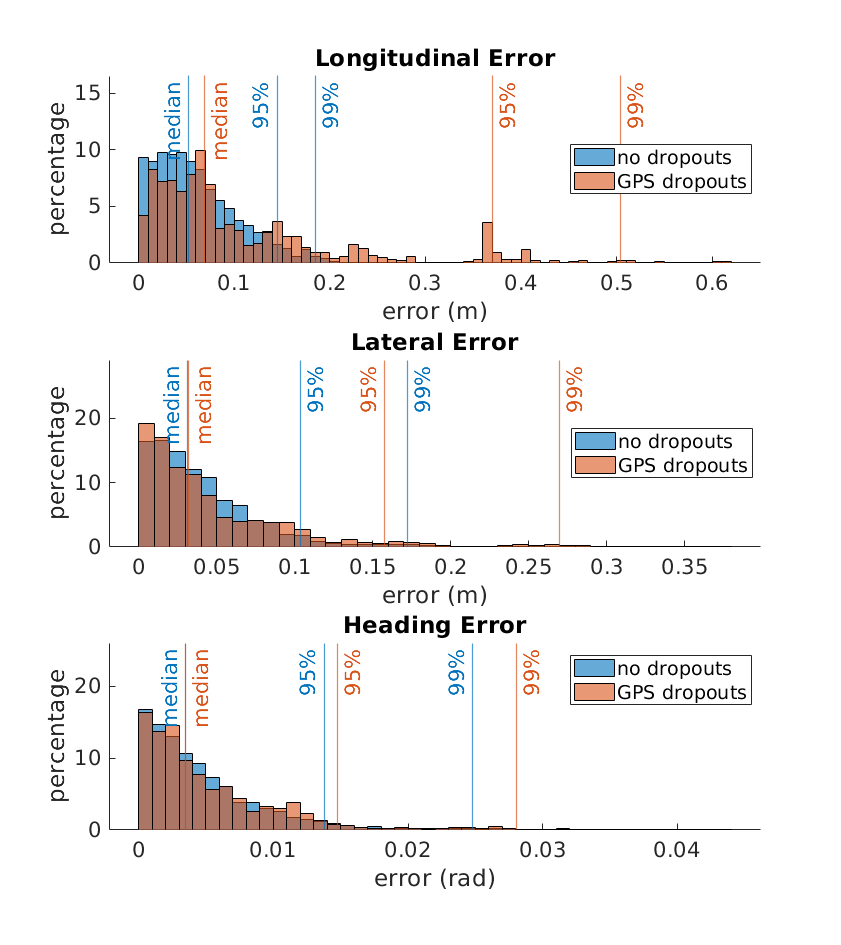}
    \vspace{-5mm}
    \caption{Histograms of longitudinal (top), lateral (middle), and heading (bottom) localization errors of Carla simulation comparing scenarios with and without GPS dropouts.}
    \label{fig:carla_hist}
    \vspace{-5mm}
\end{figure}

\begin{figure}[t]
    \centering
    \includegraphics[trim=0 5 0 0,clip,width=\columnwidth]{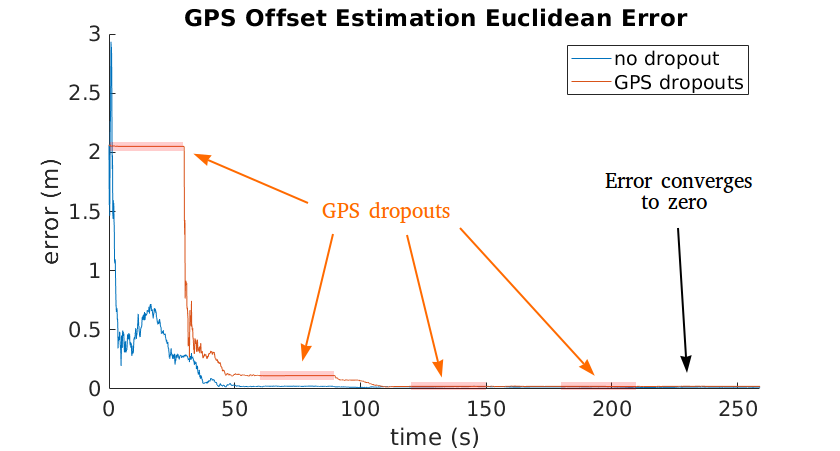}
    \vspace{-5mm}
    \caption{Euclidean error of GPS-to-map offset estimation of Carla simulation. By taking advantage of semantic cues, our localization algorithm is able to estimate the GPS measurement offset with decimetre-level accuracy even with the presence of periodic GPS dropouts.}
    \vspace{-5mm}
    \label{fig:carla_gps_offset}
\end{figure}

To evaluate the robustness of our localization algorithm, we introduced periodic GPS dropouts lasting for 30 seconds in every 60-second interval, i.e., half of the GPS measurements are lost. Under such conditions, the proposed approach is still able to achieve accurate estimation of the GPS offset as shown by the orange line in Figure~\ref{fig:carla_gps_offset}, albeit at a slower pace. Furthermore, the localization results are shown in Figure~\ref{fig:carla_hist} and summarized in Table~\ref{table:carla_results}. Compared to the scenario without any GPS dropout, we observe virtually no increase in the median lateral and heading errors largely due to frequent occurrences of lane markings, which keep the vehicle in the correct lane. This highlights the importance and effectiveness of lane markings as a crucial type of semantic cues in semantic localization. On the other hand, there is a significant decline in performance over the worst case scenario in terms of longitudinal error, where it increases from 0.185~m to 0.504~m. This can be attributed to the infrequent appearance of traffic lights, which help with longitudinal localization, when road intersections are not nearby. In this case, the vehicle can only rely on wheel odometry for relative localization during GPS dropouts, which accumulates longitudinal errors. Nevertheless, the localization accuracy is still acceptable for autonomous driving. This demonstrates the robustness of our proposed approach against frequent GPS dropouts by leveraging semantic cues.

\begin{table}[t]
\caption{Carla localization accuracy with \& without GPS dropouts}
% \vspace{-4mm}
\setlength\tabcolsep{4pt}
\begin{center}
\begin{tabular}{|c|c|c|c|c|c|c|}
    \hline
    & \multicolumn{6}{c|}{Experimental Scenarios}\\
    \cline{2-7}
    & \multicolumn{3}{c|}{No Dropouts} & \multicolumn{3}{c|}{GPS Dropouts}\\
    \cline{2-7}
    Errors & Median & 95\% & 99\% & Median & 95\% & 99\% \\
    \hline
    Longitudinal (m) & 0.053 & 0.145 & 0.185 & 0.069 & 0.370 & 0.504\\
    Lateral (m) & 0.031 & 0.104 & 0.172 & 0.032 & 0.158 & 0.270\\
    Heading (rad) & 0.004 & 0.014 & 0.025 & 0.004 & 0.015 & 0.028\\
    \hline
\end{tabular}
\vspace{-6mm}
\label{table:carla_results}
\end{center}
\end{table}

\subsection{Real-World Experiments}
Unfortunately, the vast majority of the publicly available self-driving datasets do not provide semantic maps. Those that do all lack other components necessary for our experiments. For instance, nuScenes dataset does not provide raw GPS data \cite{nuscenes2019}. Therefore, to demonstrate real-world feasibility, experiments are conducted using aUToronto's dataset collected during the SAE AutoDrive competition where the incident of uncalibrated GPS occurred \cite{doi:10.1002/rob.21958}. However, due to the lack of localization ground truth in the dataset for comparison, this will only serve as anecdotal results to verify the effectiveness of our approach. Figure~\ref{fig:gps_vs_semantic} provides a side-by-side comparison of a snapshot of the camera image overlaid with projection of the lane boundaries from the semantic map, which indicates the estimated location of the vehicle with respect to the map. Visually, we see that the GPS data is unusable on its own while our approach is able to self-calibrate for the GPS offset and has the projected lane boundaries well aligned with the lane markings to achieve accurate localization.

\section{CONCLUSION AND FUTURE WORK}
\label{sec:conclusion}
In this paper, we proposed a method capable of localizing an autonomous vehicle while self-calibrating for an offset between GPS and semantic map frames. This is achieved by using a lightweight semantic map containing locations of lane boundaries and traffic lights, which are complementary in correcting for lateral and longitudinal position of the vehicle. These semantic cues are detected via a monocular camera and integrated with GPS and wheel encoders. Our approach is evaluated using Carla simulator, which demonstrates robustness against GPS dropouts in addition to achieving decimetre-level accuracy. The real-world feasibility of our approach is also validated with a dataset collected by an vehicle with a GPS not calibrated to the semantic map. Next steps include the addition of other types of semantic cues to decrease the gap between semantic cue observations due to sparsity, as well as the closed-loop implementation of our semantic localization system on an autonomous vehicle.

\section*{Acknowledgment}

This work is funded by the Natural Sciences and Engineering Research Council of Canada (NSERC), and supported by aUToronto who provided access to their internal software and dataset.

\bibliographystyle{IEEEtran}
\bibliography{refs}

\end{document}